\documentclass{article} 
\usepackage[final]{colm2026_conference}

\usepackage{microtype}
\usepackage{hyperref}
\usepackage{url}
\usepackage{booktabs}
\usepackage{fontawesome5}

\usepackage{lineno}

\usepackage{amsthm}
\usepackage{amssymb}
\usepackage{mathtools}
\usepackage{xcolor}  
\usepackage{hyperref}
\usepackage[nameinlink,capitalize]{cleveref}
\hypersetup{colorlinks=true,linkcolor=blue,citecolor=blue,urlcolor=blue,pdfborder={0 0 0}}
\usepackage[normalem]{ulem} 
\usepackage{mathtools}

\usepackage[utf8]{inputenc} 
\usepackage[T1]{fontenc}    
\usepackage{url}            
\usepackage{booktabs}       
\usepackage{amsfonts}       
\usepackage{nicefrac}       
\usepackage{microtype}      
\usepackage{proof-at-the-end}  
\usepackage{multirow}

\usepackage{tcolorbox}
\tcbuselibrary{breakable,skins}
\usepackage{listings}
\usepackage{graphicx,wrapfig}
\usepackage{algorithm}
\usepackage{algpseudocode}
\usepackage{amsfonts}
\usepackage{url}
\usepackage{enumitem}
\usepackage{subcaption}
\usepackage{amsthm}
\usepackage{thmtools}
\usepackage{thm-restate}

\definecolor{zrcgreen}{RGB}{0,110,60}

\newcommand{\method}{HPFA}
\newcommand{\wwo}{\texttt{w/o}}

\newcommand{\cG}{{\mathcal{G}}}

\newcommand{\cK}{{\mathcal{K}}}

\newcommand{\cP}{{\mathcal{P}}}
\newcommand{\cQ}{{\mathcal{Q}}}
\newcommand{\cR}{{\mathcal{R}}}
\newcommand{\cS}{{\mathcal{S}}}
\newcommand{\cT}{{\mathcal{T}}}

\newcommand{\bc}{\begin{center}}
\newcommand{\ec}{\end{center}}

\newcommand{\bdm}{\begin{displaymath}}
\newcommand{\edm}{\end{displaymath}}

\newcommand{\beq}{\begin{equation}}
\newcommand{\eeq}{\end{equation}}

\newcommand{\bfl}{\begin{flushleft}}
\newcommand{\efl}{\end{flushleft}}

\newcommand{\bt}{\begin{tabbing}}
\newcommand{\et}{\end{tabbing}}

\newcommand{\beqn}{\begin{align}}
\newcommand{\eeqn}{\end{align}}

\newcommand{\beqs}{\begin{align*}} 
\newcommand{\eeqs}{\end{align*}}  

\tcbset{
  promptquotebase/.style={
    enhanced,
    breakable,
    skin=enhanced,
    fonttitle=\small,
    coltitle=white,
    colbacktitle=black!58,
    titlerule=0pt,
    toptitle=1.2mm,
    bottomtitle=0.8mm,
    colback=yellow!4!white,
    colframe=black!30,
    boxrule=0.55pt,
    arc=2.5pt,
    left=11pt,
    right=11pt,
    top=9pt,
    bottom=10pt,
    before skip=\medskipamount,
    after skip=\medskipamount,
    borderline west={3.2pt}{0pt}{brown!58!black},
    drop fuzzy shadow,
  },
}
\newenvironment{promptcard}[1]{%
  \begin{tcolorbox}[promptquotebase, title={\sffamily\bfseries\small #1}]%
}{%
  \end{tcolorbox}%
}
\lstdefinelanguage{promptraw}{}
\lstdefinestyle{promptsnippet}{
  language=promptraw,
  basicstyle=\ttfamily\footnotesize,
  breaklines=true,
  breakatwhitespace=false,
  columns=fullflexible,
  frame=none,
  keepspaces=true,
  showstringspaces=false,
}

\definecolor{darkblue}{rgb}{0, 0, 0.5}
\hypersetup{colorlinks=true, citecolor=darkblue, linkcolor=darkblue, urlcolor=darkblue}

\title{\method: Hypergraph-Based Paired Failure Attribution for LLM Reasoning}

\newcommand{\nus}{\textsuperscript{1}}
\newcommand{\pku}{\textsuperscript{2}}
\newcommand{\utexas}{\textsuperscript{3}}
\newcommand{\ulux}{\textsuperscript{4}}

\author{Runchuan Zhu\nus\textsuperscript{*},
Hongbin Lai\pku\textsuperscript{*},
Bowen Jiang\pku,
Junrui Zhang\nus,\\
{\bfseries Zhangheng Li\utexas,
Ostap Kilbasovych\ulux,
Junyuan Hong\nus\textsuperscript{\textdagger}}\\\\
\nus National University of Singapore,
\pku Peking University, \utexas University of Texas at Austin, \\
\ulux Univ. Bretagne Sud; Univ. of Luxembourg; EMJM in Cybersecurity (CYBERUS) \\  
\textsuperscript{*}Equal contribution. \textsuperscript{\textdagger}Corresponding author.\\
\faEnvelope~\texttt{zhurunchuan0828@gmail.com; jyhong@nus.edu.sg}\\
\faGlobe~Code: \url{https://github.com/costa-nus/COLM26_HPFA}
}

\begin{document}

\ifcolmsubmission
\linenumbers
\fi

\maketitle

\begin{abstract}

Reflection is a powerful mechanism for LLM reasoning, yet its effectiveness hinges on accurately attributing failures to specific reasoning steps, a capability that current models notably lack. Existing failure attribution methods either require expensive step-by-step counterfactual testing that scales poorly with trajectory length, or treat reasoning traces as flat sequences that ignore the inherent non-linear logical dependencies. We propose a hypergraph-based paired failure attribution (\method) framework that attributes the failure root cause by comparing the hyperedges of the targeted failure reasoning path against a reference successful path. By reducing the search space, our method efficiently localizes root causes and enables scalable synthesis of attribution data for training a lightweight attributor model via supervised fine-tuning and reinforcement learning. Experiments on mathematical reasoning and agentic coding tasks demonstrate that \method{} can dramatically increase attribution accuracy and efficiency, and the trained attributor consistently improves reasoning accuracy at test time, outperforming baselines that lack graph structure or paired analysis.

\end{abstract}

\section{Introduction}


Large language models (LLMs) have demonstrated remarkable capabilities in complex reasoning tasks, from mathematical problem solving to multi-step agent decision-making~\citep{wei2022chain,openai2024o1}. A key enabler of these advances is \emph{reflective reasoning}: given execution outcomes, an LLM revisits the reasoning trace, identifies potential mistakes, and iteratively refines its output~\citep{shinn2023reflexion,madaan2023selfrefine,yin2025llm,ding2026scaling}. 
The capability is fundamental to optimize LLM systems~\citep{yuksekgonul2024textgrad,yin2025llmautodiff}, build complex compound agentic systems~\citep{agrawal2025gepa}, or optimize formal programs for robots~\citep{yang2025ad}.

\begin{figure}[t]
\vspace{-0.1in}
\centering
\includegraphics[width=0.95\linewidth]{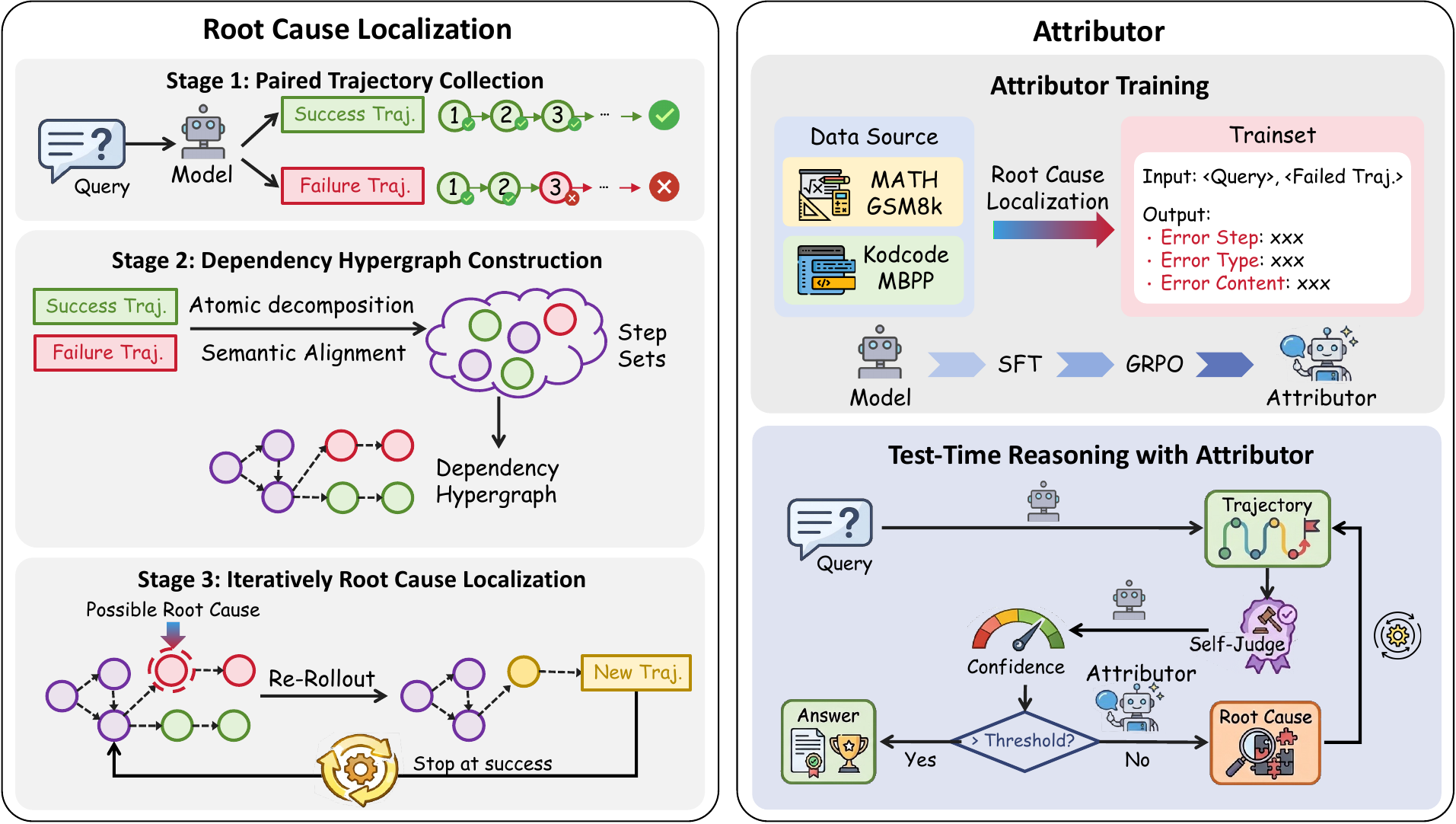}
\caption{Through paired reasoning trajectory generation and hypergraph construction, \method{} localizes root causes in failure trajectories. Attributed trajectories are used to train attributor models, which can improve reflecive reasoning.}
\label{fig:method-overview}
\vspace{-0.2in}
\end{figure}

However, effective reflective reasoning critically depends on \textbf{failure attribution}: \emph{localizing the intermediate step that is the root cause of downstream failures}. Recent evidence shows that LLMs are surprisingly poor at locating their own reasoning errors, even though they can correct mistakes once the error location is provided~\citep{tyen2024llms}.
This limitation makes failed reasoning hard to explain and debug, especially in high-stakes settings where understanding the failure cause is essential for trust and safety~\citep{qu2026medieval,braitsch2026information}.
\citet{zhang2025which} defined failure attribution in multi-agent systems to identify the failure cause without relying on the model's own introspective judgment. They also introduced the Who\&When benchmark, showing that existing LLMs have limited ability to localize failure points.



A growing body of work has attempted to tackle this problem. One line uses taxonomy-guided prompting to analyze a failed trajectory, identify a critical step, and generate corrective feedback~\citep{zhu2025where}. Another constructs attribution data through counterfactual replay and programmed fault injection, then trains a dedicated attributor to predict the decisive error step directly from the failed trajectory~\citep{zhang2025agentracer}.
Training a dedicated attributor is particularly promising because, once trained, it can localize errors directly from a failed trajectory without repeated replay at inference time.
We therefore focus on constructing scalable attribution data for training such an attributor.


However, reliably attributing complex, long-chain trajectories remains challenging for two reasons.
\textbf{(1)~Expensive data synthesis.}
Step-by-step counterfactual labeling repairs each candidate step and re-executes the downstream trajectory, so its cost scales quickly with trajectory length~\citep{zhang2025agentracer}.
Tasks such as mathematical proofs and agent planning containing dozens of steps could suffer from expensive large-scale synthesis.
\textbf{(2)~Non-linear dependencies.}
Prior methods represent trajectories as linear sequences, overlooking non-local and higher-order dependencies across steps.
For example, in a multi-hop proof, a later conclusion may depend jointly on several earlier parallel math derivations that are far apart in the reasoning trace.
This obscures how early errors propagate and makes upstream root causes difficult to identify.


We address the challenges by the \emph{Hypergraph-based Paired Failure Attribution} (\method{} in \cref{fig:method-overview}) framework that accelerates synthesis of attributed reasoning trajectories by two major steps.
\textbf{(1)~Paired trajectories for counterfactual search.} For each query, a successful trajectory provides a valid reference for the failed trajectory, which reduces search to divergences instead of probing every step.
\textbf{(2)~Dependency hypergraphs capture non-linear error propagation.} We represent atomic reasoning steps as nodes and their higher-order dependencies as hyperedges, allowing attribution to follow non-linear dependency paths back to upstream root causes.
Our key contributions are as follows:
\begin{itemize}[leftmargin=*]
    \item \textbf{Efficient Attribution-Data Synthesis.} We introduce \method{}, a paired-trajectory and dependency-hypergraph pipeline that synthesizes root-cause supervision while reducing counterfactual search. Across mathematical reasoning, agentic coding, and privacy tasks, \method{} achieves a better localization cost--success trade-off than existing baselines.
    \item \textbf{Better Attribution Model.} We train a stronger attribution model on \method{} generated data to attribute failure for reasoning path. We validate its effectiveness in reflection scenarios on mathematical reasoning and agentic coding tasks.
\end{itemize}

\section{Related Work}

\textbf{Failure Attribution in LLM Reasoning.}
Failure attribution plays an important role in reflective reasoning. It identifies the root cause in a failed reasoning trajectory and provides feedback signals for reflection or RL training~\citep{lightman2023verify,wang2024mathshepherd}. Self reflection comprises error detection, error localization, and reasoning revision, with localization connecting failure recognition to targeted revision~\citep{samanta2026structure}. As LLMs struggle to locate logical mistakes, providing failure causes can improve revision accuracy~\citep{tyen2024llms,li2024evaluating}. (1) Methods for acquiring this capability span three paradigms. Prompting methods elicit self-rethinking, error type analysis, or first error localization, but remain limited by the model's own verification ability~\citep{tong2024learn,li2024evaluating,samanta2026structure}. (2) Supervised fine-tuning (SFT) methods use erroneous traces and mistake annotations to train reasoners that learn from mistakes, while dedicated mistake locators can be trained from step labels~\citep{tong2024learn,tyen2024llms}. (3) In reinforcement learning, process reward models (PRMs) play a similar role. PRMs provide stepwise critics that localize the first error~\citep{zha2025rltango,zhang2026stride}, which can assign credit to correct prefixes and erroneous suffixes~\citep{liu2026save}. 

\textbf{Failure Attribution in Agentic Systems.}
Different from LLMs, agentic systems are compound systems where failures can stem from reasoning, memory, tool execution and agent communication~\citep{ji2024testing,ning2024defining,zhu2025where,cemri2025why}, as upstream errors can propagate across components and subsequent decisions, thereby separating observed failures from their responsible components and originating steps~\citep{zhu2025where,zhang2025agentracer,zhang2025which}.
Early approaches frame agent failure analysis as specification-based verification. For example, PDoctor checks plans against constraints and test oracles~\citep{ji2024testing}, whereas VeriLA verifies outputs against human-defined criteria~\citep{sung2025verila}. These methods can detect known failure patterns but cannot determine which component caused an end-to-end failure or at which step the failure first occurred. Who\&When marks the shift from specification-based verification to automated failure attribution~\citep{zhang2025which}.
Following this shift, AgentDebug uses a taxonomy to retrospectively diagnose full trajectories and generate corrective feedback~\citep{zhu2025where}, whereas AgenTracer trains a lightweight attributor for agent- and step-level localization using labels derived from counterfactual replay and fault injection~\citep{zhang2025agentracer}.and critical step~\citep{zhang2025agentracer}. 

Despite the success of the aforementioned methods in reasoning or agentic systems, there are two remaining challenges: (1) Such a labeling method is costly with repeated interventions and executions. (2) Existing methods do not explicitly model complex dependencies among agents, steps, and environmental factors. Our work aims to address the two challenges in failure attribution for LLM reasoning and agentic systems.


\section{Methodology}
\label{sec:Methodology}

\noindent\textbf{Problem Formulation.}
Our goal is to pinpoint a \emph{root cause} in the trajectory of a failed rollout: a step that is chiefly responsible for the error, in the sense that replacing it could yield a successful outcome. We formalize this as follows.
Given a problem instance (e.g., a query), a \emph{reasoning trajectory} is $\cT=(s_1,\ldots,s_n,y)$, which includes intermediate \emph{steps} $s_i$ (e.g., propositions, subgoals, or tool outputs) and final prediction $y$. Let $\mathbb{1}(\cT)\in\{0,1\}$ indicate whether $\cT$ is correct (e.g., final answer vs.\ ground-truth $y^\star$).
For a failed rollout with $\mathbb{1}(\cT_{\mathrm{fail}})=0$, \emph{Root-cause attribution} looks for the earliest step whose correction can recover the final outcome. For an index $k$ and a same-granularity replacement $s_k'$ of step $s_k$, define $\mathrm{R}(\cT,k,s_k')$ as the \emph{new} trajectory obtained by keeping the prefix $(s_1,\ldots,s_{k-1})$ fixed, substituting $s_k\leftarrow s_k'$, and then re-sampling all downstream steps $(s_{k+1},\ldots,s_n,y)$ under a fixed rollout protocol until completion. We call $s_k$ a \emph{root cause} if
 \[
 k=\min\left\{j:\exists s_j'\ \text{s.t.}\ \mathbb{1}\bigl(\mathrm{R}(\cT_{\mathrm{fail}},j,s_j')\bigr)=1\right\}.
 \]
 This earliest-recoverable-step convention handles cases where several downstream steps could be repaired: they are treated as consequences of the first recoverable divergence (root cause) rather than separate primary causes.

\subsection{Failure Localization via Hypergraph-based Paired Failure Attribution (\method{})}
\label{sec:root-cause-localization}

A direct way to localize root cause is to probe \emph{every} step of a failed trace with counterfactual rollouts $\mathrm{R}$, which scales poorly with trace length and perturbation budget, especially on long-horizon reasoning or agentic tasks where each probe is expensive. 
To address the challenge, we propose \emph{Hypergraph-based Paired Failure Attribution} (\method{}) in \cref{alg:root-cause-localization}, which reduces the search space by providing a successful reasoning reference, and guiding the search in non-linear dependency via a hypergraph.


\noindent\textbf{Step 1: Paired-trajectory generation.}
To narrow down the search space of failure causes, we directly sample additional reasoning trajectories from the same model and find a successful one as a reference.
Prior work on \emph{test-time scaling} has shown that repeatedly sampling from the same model can surface at least one correct solution far more effectively~\citep{snell2024scaling,chen2021evaluating,cobbe2021training}.
As a result, the search complexity is proportional to $1/p$ instead of reasoning steps, where $p$ is the probability of getting the successful trace in one shot.
In \method{}, we introduce \emph{trajectory pairing}: once a successful rollout $\cT^{+}$ emerges on a query, comparing it to a failure $\cT^{-}$ supplies a reference of valid intermediate structure and narrows root-cause search to genuine divergences instead of probing every possible step in the trace.
For each query $\cQ$, we roll out the target LLM or agent under a fixed protocol until both outcomes appear or we exhaust a per-query budget of $\cK$ attempts. If only one outcome appears, we drop $\cQ$. We keep one success-failure pair $(\cT^{+},\cT^{-})$ with $\mathbb{1}(\cT^{+})=1$ and $\mathbb{1}(\cT^{-})=0$. When several qualify, we keep the pair with the closest length.
Trajectory-sampling prompt templates (coding) are listed in \cref{app:prompts_en}.

\noindent\textbf{Step 2: Dependency hypergraph construction.}
Even with reference, it is still challenging to find the failure location by comparison, because the step-to-step \emph{dependencies} in reasoning could be non-linear, complicating the search.
For example, later content relies on earlier and non-consecutive propositions, tool observations, or subgoals. Rather than treating the trace as a flat sequence, we summarize these dependencies in a structured graph-like object $\cG=(\mathcal{V},\mathcal{E}_h)$ over aligned abstract steps, so that we can walk backward from the final prediction toward regions where a root cause is plausible. We instantiate $\cG$ as a \emph{hypergraph}.
Given $(\cT^{+},\cT^{-})$, we segment each trace into \textbf{atomic} steps $(s^{+}_1,\ldots,s^{+}_{n^{+}})$ and $(s^{-}_1,\ldots,s^{-}_{n^{-}})$ (e.g., one arithmetic move, tool call, or subgoal). Semantic similarity and clustering induce an equivalence $\sim$ on the union of atomic steps. We write $[s]_{\sim}$ for the class of $s$, so that
\[
\begin{aligned}
  s \sim s' \quad &\Longleftrightarrow \quad [s]_{\sim} = [s']_{\sim}, \\
  \mathcal{V} &= \bigl\{\, [s]_{\sim} : s \in \{s^{+}_1,\ldots,s^{+}_{n^{+}}, s^{-}_1,\ldots,s^{-}_{n^{-}}\} \,\bigr\}.
\end{aligned}
\]
The elements of $\mathcal{V}$ are the \emph{abstract steps} used as hypergraph vertices.
We merge two steps only when both traces explicitly express the same reasoning operation or code intent; otherwise, they remain distinct.
For each trajectory we estimate prerequisites $\{s_i\}_{i\in \cP(j)}$ for $s_j$ and encode them as hyperedges over $\mathcal{V}$, yielding $\cG=(\mathcal{V},\mathcal{E}_h)$. Let $v_{\mathrm{out}}^{+}$ and $v_{\mathrm{out}}^{-}$ denote the vertices for the final predictions on $\cT^{+}$ and $\cT^{-}$.
We construct $\cG$ with an LLM using the hypergraph construction template in \cref{app:hypergraph_prompt_en} and validate the reliability of the hypergraph in \cref{app:hypergraph-validation}.

\noindent\textbf{Step 3: Iterative root-cause localization.}
\label{sec:iterative-localization}
With the constructed hypergraph, we search for the root cause of the reasoning failure.
We denote the abstract vertices used by $\cT^{+}$ and $\cT^{-}$ as $\mathcal{V}^{+}$ and $\mathcal{V}^{-}$. Starting from $v_{\mathrm{out}}^{+}$, walk backward along dependencies on $\cT^{+}$ (reverse step order; fixed tie-breaking at branches). The first vertex in this walk that lies in $\mathcal{V}^{+}\cap\mathcal{V}^{-}$ is the \emph{meeting} vertex $v^{\star}$:
\[
  v^{\star}\in \mathcal{V}^{+}\cap\mathcal{V}^{-}.
\]
This $v^{\star}$ need not coincide with an undirected lowest common ancestor (LCA) of $v_{\mathrm{out}}^{+}$ and $v_{\mathrm{out}}^{-}$, because the walk is directed along $\cT^{+}$. On $\cT^{-}$, successors of $v^{\star}$ not aligned to $\cT^{+}$ under $\sim$ form the candidate set $\mathcal{C}$. For each $v\in\mathcal{C}$, map $v$ to an index $k$ on $\cT^{-}$, substitute a repaired span $s'$, and test $\mathbb{1}(\mathrm{R}(\cT^{-},k,s'))$: any verifying candidate is the \emph{root cause}. Otherwise re-anchor at $v^{\star}$, walk backward to a new meeting vertex and $\mathcal{C}$, and repeat. The search ends when a repair succeeds, the shared prefix is exhausted, or we fall back to the earliest unresolved divergence.

\begin{algorithm}[t]
\caption{Hypergraph-based Paired Failure Attribution (\method{})}
\label{alg:root-cause-localization}
\begin{algorithmic}[1]
\Require Query $\cQ$, correctness oracle $\mathbb{1}(\cdot)$, rollout cap $\cK$, counterfactual protocol $\mathrm{R}$
\Ensure Root step index $k$ on $\cT^{-}$, or \textsc{Fail}
\State Roll out until both outcomes occur or $\cK$ attempts are exhausted
\If{no success--failure pair} \Return \textsc{Fail} \EndIf
\State $(\cT^{+},\cT^{-}) \gets$ chosen pair; \quad $(\mathcal{V},\cG,v_{\mathrm{out}}^{\pm}) \gets \textsc{AlignGraph}(\cT^{+},\cT^{-})$
\State $\textit{anchor} \gets v_{\mathrm{out}}^{+}$
\While{\textbf{true}}
  \State $v^{\star} \gets \textsc{Meet}(\textit{anchor},\cT^{+})$ \Comment{meeting vertex along $\cT^{+}$}
  \State $\mathcal{C} \gets \textsc{Cands}(v^{\star},\cT^{-})$ \Comment{candidates on $\cT^{-}$}
  \For{$v\in\mathcal{C}$}
    \State $\tilde{\cT}\gets \mathrm{R}(\cT^{-},\text{idx}(v),s')$ \Comment{counterfactual rollout}
    \If{$\mathbb{1}(\tilde{\cT})=1$} \Return $\text{idx}(v)$ \EndIf
  \EndFor
  \If{$v^{\star}=\textit{anchor}$} \Return \textsc{Fail} \EndIf
  \State $\textit{anchor} \gets v^{\star}$ \Comment{re-anchor}
\EndWhile
\end{algorithmic}
\end{algorithm}

\subsection{Boosting Reasoning by Test-time Attributors}
\label{sec:attributor}

To boost the reasoning at test time where ground truths or references are not available, we train a lightweight \emph{attributor} $\pi_\phi$ that maps $(\cQ,\cT_{\mathrm{fail}})$ to a structured diagnosis: the erroneous step, its error type, and a concise description of the mistaken content. Supervision is harvested by \method{}: each training tuple includes the failure trace $\cT^{-}$, the verified root index $k$, a short rationale, and optionally the counterfactual span used when a repair rollout succeeds, i.e., $\mathbb{1}(\mathrm{R}(\cT^{-},k,s'))=1$.

\noindent\textbf{Training attributor.}
The training follows a \emph{two-stage} schedule that separates imitation from on-policy refinement. \textbf{(1)~SFT.} We fit $\pi_\phi$ \emph{offline} with the standard causal language-modeling objective. Prompts concatenate the task instruction, $\cQ$, and the indexed failure trace; targets are templated diagnoses (step, error type, mistaken content). For math, paired and negative-only root-cause prompts (\cref{app:paired_rca_prompt_en,app:failed_only_rca_en}) match the \texttt{root\_cause} tag schema used by the format reward. \textbf{(2)~GRPO.} We use the final model to initiate the Group Relative Policy Optimization (GRPO)~\citep{deepseekmath2024} on failures rolled out \emph{on-policy} from the target model. Thus, the training distribution can match deployment-time errors. 

For each $(\cQ,\cT_{\mathrm{fail}})$ we sample a group of attributions from $\pi_\phi$, score each with the sequence-level reward below, normalize within-group rewards to advantages (zero mean, unit variance, with a small stability constant), and update using clipped policy ratios and KL regularization to $\pi_{\mathrm{ref}}$.
Each sampled attribution $o_i$ receives a scalar reward decomposed into a \emph{format} term and a \emph{rollout} term:
\begin{align}
  \cR_i &= \cR_i^{\mathrm{fmt}} + \lambda\, \cR_i^{\mathrm{rollout}}, \quad \lambda>0, \ \cR_i^{\mathrm{rollout}} = \mathbb{1}(\tilde{\cT}_i).
\end{align}
The format reward $\cR_i^{\mathrm{fmt}}$ verifies that $o_i$ parses under our diagnosis template: the reported step index is valid, required fields are present, and delimiters match the SFT target schema so downstream repair can consume $o_i$ without heuristics. The rollout reward $\cR_i^{\mathrm{rollout}}$ tests \emph{actionability}: we condition the base LLM or agent on $(\cQ,\cT_{\mathrm{fail}},o_i)$ and draw a new trajectory $\tilde{\cT}_i$; success is indicated by $\mathbb{1}(\tilde{\cT}_i)$. Together, $\cR_i$ favors diagnoses that are both well-formed and sufficient to elicit a correct re-rollout, not merely fluent text.

\noindent\textbf{Reflective reasoning with attributor.}
In \cref{alg:test-time-reflection}, we incorporate the trained attributor to enhance the reasoning at test time.
Different from failure localization, the attributor here is not aware of the ground truth or the successful path.
At test time, the system sees a single backbone rollout for $\cQ$. We call $\pi_\phi$ on $(\cQ,\cT)$, without graph or $\cT^{+}$. 
We design the algorithm to (1) infer the correctness of the answer and (2) attribute and correct the trajectories upon the assumed wrong answer.
We use verbal confidence score $\cS(\cT)$ to decide whether to trigger the attribution.
If $\cS(\cT)<\tau$, the LLM passes its reasoning trajectory to the attributor, and then regenerates the trajectory based on the output of the attribution model. The corresponding prompt is provided in \cref{app:rereason_prompt_en}.

\begin{algorithm}[t]
\caption{Reasoning with attributor}
\label{alg:test-time-reflection}
\begin{algorithmic}[1]
\Require $\cQ,\pi,\pi_\phi,\cS,\tau,B$
\Ensure $(\cT,y)$
\State $(\cT,y) \gets \textsc{Rollout}(\pi,\cQ)$
\For{$t = 1,\ldots,B$}
  \If{$\cS(\cT)\ge\tau$} \Return $(\cT,y)$ \EndIf
  \State $(\cT',y') \gets \textsc{Repair}\bigl(\pi,\cQ,\cT,\pi_\phi(\cQ,\cT)\bigr)$
  \If{$\cS(\cT')\ge\tau$ \textbf{or} negligible $\cS(\cT')\!-\!\cS(\cT)$} \Return $(\cT',y')$ \EndIf
  \State $(\cT,y)\gets(\cT',y')$
\EndFor
\Return $(\cT,y)$
\end{algorithmic}
\end{algorithm}

\section{Experiments}
\label{sec:experiments}

We organize evaluation along two components: (i) failure localization with \method{}, and (ii) a trained attributor used for reflective reasoning.

\subsection{Failure Localization}
\label{sec:exp-localization}

We evaluate \method{}\ in the paired-trajectory setting: for each query we contrast a successful trajectory $\cT^{+}$ with a failed one $\cT^{-}$, build a dependency hypergraph, and use counterfactual rollouts to verify the attributed root step (\cref{sec:root-cause-localization}).
The subsections below specify \emph{where} we test (datasets), \emph{how} we obtain pairs and baselines, \emph{what} we measure, and then interpret accuracy, convergence, and cost together.

\noindent\textbf{Datasets.}
We evaluate \method{} in three representative settings: mathematical reasoning, agentic coding, and a privacy scenario.
\emph{(i)~Mathematical reasoning:} We study multi-step mathematical problem solving on MATH~\citep{hendrycks2021measuring} and GSM8K~\citep{cobbe2021training}, using their official training splits.
\emph{(ii)~Agentic coding:} We study coding agents on KodCode~\citep{xu2025kodcode} and MBPP~\citep{austin2021program}. We use the KodCode training split and a $3{:}1$ train--evaluation split on MBPP.
\emph{(iii)~Privacy scenario:} We study reasoning failures caused by privacy-preserving de-identification. To construct the dataset, we use PrivateAI to de-identify CoQA-based inputs~\citep{reddy2019coqa}. More details are in \cref{sec:app-privacy-localization}.

\noindent\textbf{Trajectory preparation.}
We sample trajectories with Step~1 of \method{}\ (\emph{paired-trajectory generation}; \cref{sec:root-cause-localization}): for each query, we roll out the target model as in \cref{alg:root-cause-localization} until both a success and a failure are observed or the per-query budget is exhausted, and keep one pair $(\cT^{+},\cT^{-})$ with $\mathbb{1}(\cT^{+})=1$ and $\mathbb{1}(\cT^{-})=0$.
\cref{tab:traj-stats} summarizes the number of trajectory pairs and successful repair traces for each dataset. The successful repair traces are used to train the attribution model described in \cref{sec:exp-attributor}.

\cref{tab:traj-stats} summarizes the number of trajectory pairs and successful repair traces for each dataset. The successful repair traces will be used to train the attribution models.

\begin{table}[t]
\centering
\caption{The number of trajectories collected for training attributors.}
\label{tab:traj-stats}
\small
\begin{tabular}{@{}lcccc@{}}
\toprule
 & \textbf{MATH} & \textbf{GSM8K} & \textbf{KodCode} & \textbf{MBPP} \\
\midrule
Success--failure pairs & 884 & 199 & 853 & 479 \\
Repair-success traces (Trainset) & 587 & 155 & 473 & 112 \\
\bottomrule
\end{tabular}
\end{table}

\noindent\textbf{Baselines.}
We compare two failure attribution baselines using counterfactual testing and including an ablation without the hypergraph.
(1)~\textbf{AgentDebug}: introduces error taxonomy for identifying root cause~\citep{zhu2025where}.
(2)~\textbf{AgenTracer}: trains a lightweight attributor to identify root cause~\citep{zhang2025agentracer}.
(3)~\textbf{Ours \wwo{} graph}: same paired trajectories and counterfactual verification as our full method, but without hypergraph-guided candidate extraction.

\noindent\textbf{Implementation.}
For math and coding, we use Qwen3-8B for both trajectory generation and dependency-hypergraph construction, with temperature $0.7$ and a maximum generation length of 8192 tokens.
We set the per-query paired rollout budget to $\cK=8$, and counterfactual repair rollouts use the same decoding temperature as trajectory sampling.
For privacy, all LLM-based components use GPT-4o; complete implementation details are provided in \cref{sec:app-privacy-localization}.

\begin{table}[t]
\centering
\caption{Rollout-verified success rate (\%) at the final (5th) refinement step of the localization loop. A method is credited on a failure trace if and only if substituting its localized step with a repaired span flips the outcome under the shared oracle $\mathbb{1}(\cdot)$ (the same oracle used by \emph{all} methods, including baselines).}
\label{tab:efficiency-attribution}
\small
\setlength{\tabcolsep}{4pt}
\begin{tabular}{@{}lccccc@{}}
\toprule
 & \multicolumn{2}{c}{\textbf{Math}} & \textbf{Privacy} & \multicolumn{2}{c}{\textbf{Coding}} \\
\cmidrule(lr){2-3} \cmidrule(lr){4-4} \cmidrule(lr){5-6}
\textbf{Method} & MATH500 & GSM8K & CoQA & KodCode & MBPP \\
\midrule
AgentDebug & 30.9 & 37.6 & 20.4 & 16.8 & 5.8 \\
AgenTracer & 27.3 & 46.2 & 22.2 & 11.5 & 5.0 \\
\midrule
Ours \wwo{} graph & 54.9 & \textbf{81.4} & 42.6 & 34.8 & 11.9 \\
\textbf{Ours} & \textbf{64.6} & 76.9 & \textbf{53.7} & \textbf{53.4} & \textbf{23.0} \\
\bottomrule
\end{tabular}
\end{table}

\noindent\textbf{Main results.}
In \cref{tab:efficiency-attribution}, we report the \emph{rollout-verified success rate} (\%): the fraction of failure traces for which repairing a method's localized step $k$ yields a successful rollout, i.e.\ $\mathbb{1}(\mathrm{R}(\cT^{-},k,s'))=1$. In the privacy setting, a rollout must additionally satisfy the no-leakage constraint described in \cref{sec:app-privacy-localization}. \textbf{(1) Versus baselines.} At the fifth refinement step, \textbf{Ours} outperforms both \textbf{AgentDebug} and \textbf{AgenTracer} on all five benchmarks. Relative to the stronger baseline on each dataset, \method{} achieves improvements of over 30\% on MATH500, GSM8K, CoQA, and KodCode, and an improvement of over 17.2\% on MBPP. \textbf{(2) Versus Ours \wwo{} graph.} Compared with \textbf{Ours \wwo{} graph}, the full method improves the success rate by $9.7$\% points on MATH500, $11.1$\% on CoQA, $18.6$\% on KodCode, and $11.1$\% on MBPP. Overall, paired success--failure trajectories provide broad gains, while hypergraph-guided search yields additional improvements on four of the five benchmarks.

\textbf{When does the hypergraph help?}
Unlike the other datasets, GSM8K does not exhibit a gain from the hypergraph.
To explain this exception, we relate average trajectory length to the marginal gain of \textbf{Ours} over \textbf{Ours \wwo{} graph} across 2,415 paired failure traces from four datasets.
As shown in \cref{tab:hypergraph-length}, GSM8K has the shortest trajectories, averaging 8.2 steps, and shows a negative marginal gain, resulting in a 4.5\% performance decrease. In contrast, the marginal gain becomes positive on MATH with 12.4 steps and a 9.7\% performance gain, MBPP with 14.0 steps and an 11.1\% performance gain, and KodCode with 18.2 steps and an 18.6\% performance gain.
This trend suggests that paired trajectories suffice for short traces with few candidate root steps, whereas hypergraph-guided pruning becomes more useful as trajectory length and dependency complexity increase.

\begin{table}[t]
\centering
\caption{Trajectory length and marginal hypergraph gain. \emph{Traj. Length $\pm$ SD} is the mean $\pm$ SD number of atomic nodes per trajectory; \emph{$\Delta$ gain} is \textbf{Ours} minus \textbf{Ours \wwo{} graph} in rollout-verified success rate.}
\label{tab:hypergraph-length}
\small
\begin{tabular}{@{}lcccc@{}}
\toprule
\textbf{Metric} & \textbf{GSM8K} & \textbf{MATH} & \textbf{MBPP} & \textbf{KodCode} \\
\midrule
Traj. Length $\pm$ SD & $8.2 \pm 3.2$ & $12.4 \pm 5.8$ & $14.0 \pm 3.2$ & $18.2 \pm 5.5$ \\
$\Delta$ gain (\%) & $-4.5$ & $+9.7$ & $+11.1$ & $+18.6$ \\
\bottomrule
\end{tabular}
\end{table}

\begin{figure}[t]
\centering
\includegraphics[width=0.8\textwidth]{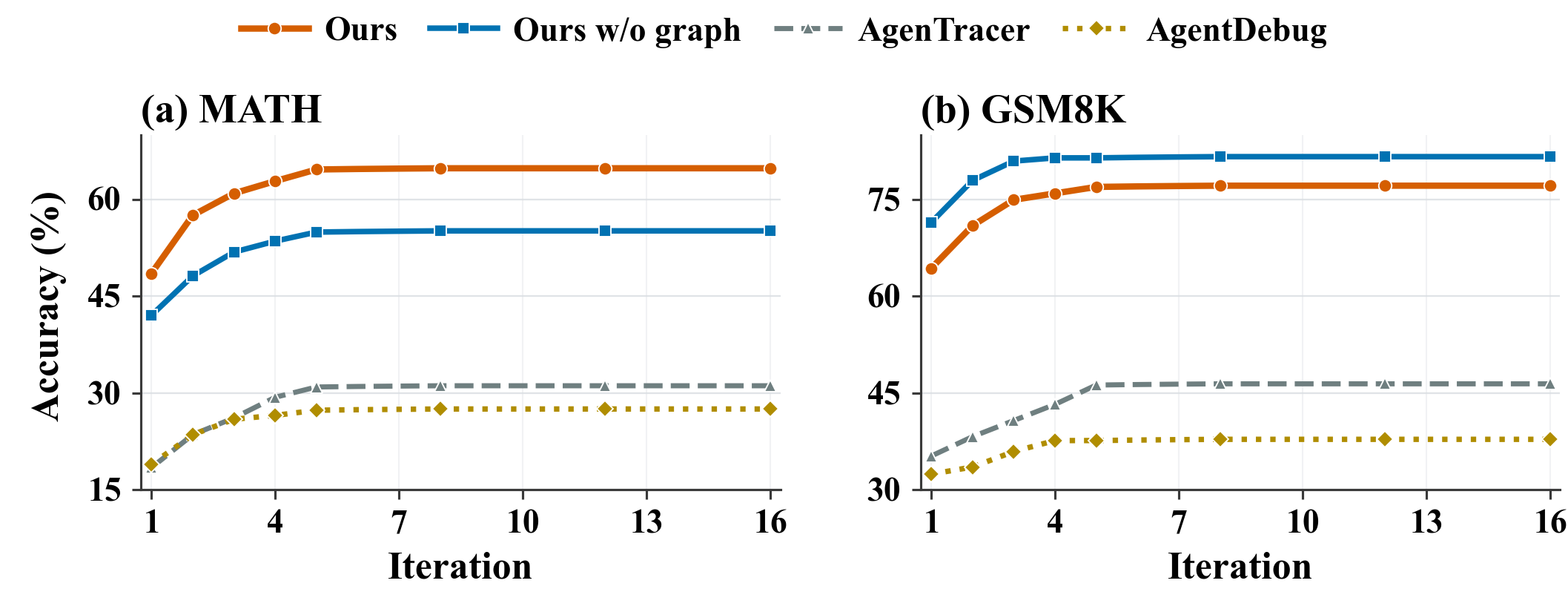}
\vspace{-0.1in}
\caption{Convergence over iterations of attribution; Our method has a similar convergence rate, where the iteration accuracy stabilizes from iteration~5 onward.}
\label{fig:convergence}
\end{figure}

\noindent\textbf{Convergence Analysis.}
Figure~\ref{fig:convergence} runs the same procedure for 16 outer rounds on the split underlying \cref{tab:efficiency-attribution}.
Accuracy rises quickly in early rounds, and \textbf{Ours} and both baselines are \emph{almost converged} from iteration~5 onward. As curves change little between rounds~5--16, we take the fifth refinement step as the representative operating point in \cref{tab:efficiency-attribution}.

\begin{figure}[t]
\centering
\includegraphics[width=0.8\textwidth]{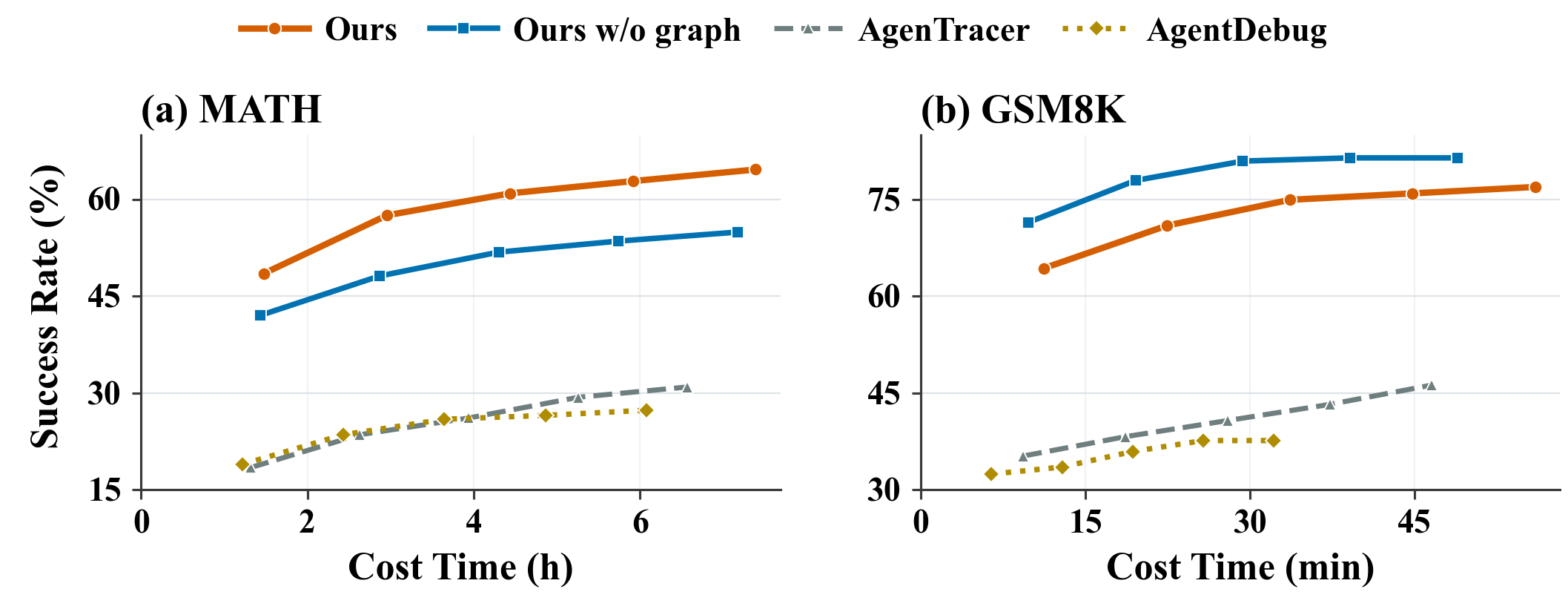}
\vspace{-0.1in}
\caption{Failure-localization success rate versus cumulative wall-clock time on (a) MATH and (b) GSM8K. Markers from left to right denote refinement iterations 1--5; additional iterations increase runtime and the chance of successful localization. Our method achieves a better cost--success trade-off than AgentDebug and AgenTracer.}
\label{fig:cost-time}
\end{figure}

\noindent\textbf{Cost vs.\ accuracy.}
Figure~\ref{fig:cost-time} shows the trade-off between computation cost and success rates over five refinement iterations. At comparable runtime, \textbf{Ours} achieves a higher localization success rate than \textbf{AgentDebug} and \textbf{AgenTracer} at similar cost. Paired success--failure trajectories focus counterfactual checks on meaningful divergences, while hypergraph-guided pruning reduces candidate root steps on complex traces, together avoiding redundant rollouts and full-trajectory rereading.

\subsection{Test-time inference with an attributor}
\label{sec:exp-attributor}

Whereas the failure localization block above uses paired trajectories $(\cT^{+},\cT^{-})$ at attribution time, here we evaluate the attributor $\pi_\phi$ in reflective reasoning.

\textbf{Datasets.}
We evaluate the attributor on MATH~\citep{hendrycks2021measuring}, GSM8K~\citep{cobbe2021training}, KodCode~\citep{xu2025kodcode}, and MBPP~\citep{austin2021program}.
For MATH, we train on the official training split and evaluate on MATH500.
For GSM8K and KodCode, we use the official training and test splits; for MBPP, we use a random $3{:}1$ train--evaluation split.
We run \method{} on each training split to synthesize attribution examples containing a failure trace $\cT^{-}$, its root step, rationale, and, when available, counterfactual span; these examples provide SFT targets and GRPO rollouts (\cref{tab:traj-stats}).
For attribution analysis (\cref{tab:attribution-rescue}), we additionally evaluate on AIME 2024~\citep{aime2024} and LogiQA~2.0~\citep{liu2023logiqa2}. AIME uses Pass@1 accuracy to identify failures and Pass@16 accuracy to evaluate repair, while LogiQA~2.0 tests non-mathematical logical reasoning.

\textbf{Baselines.}
(1) \textbf{Vanilla prompting}: single-shot decoding without explicit failure attribution; (2) \textbf{Self-reflection}:  the model names a faulty step and reflects using its own attribution (no hypergraph-derived supervision); (3) \textbf{AgenTracer}: Using the same datasets as ours, but adopting the data generation and training approach from AgenTracer~\citep{zhang2025agentracer}; (4) \textbf{Ours (SFT)}: Using data generated by \method{}, but only applying the SFT stage described in \cref{sec:attributor} while \textbf{Ours (SFT + GRPO)} performs both SFT and GRPO.

\textbf{Implementation.}
The base model for math and coding experiments is Qwen3-8B with temperature $0$, no-think mode, and a maximum generation length of 8192 tokens unless stated otherwise.
We train the attributor with supervised fine-tuning on four NVIDIA A100 (80\,GB) GPUs with batch size 16, AdamW, learning rate $5{\times}10^{-5}$, and 5 epochs, then optimize with GRPO (group size 4, 4 groups per batch, learning rate $2{\times}10^{-6}$).

\begin{table}[h]
\centering
\caption{Task accuracy (\%) using different attribution or self-reflection methods.}
\label{tab:self-reflection-main}
\vspace{-0.1in}
\small
\begin{tabular}{@{}lcccc@{}}
\toprule
\textbf{Method} & \textbf{MATH500} & \textbf{GSM8K} & \textbf{KodCode} & \textbf{MBPP} \\
\midrule
Vanilla prompting & 79.4 & 92.6 & 43.9 & 46.5 \\
Self-reflection & 81.8 & 93.2 & 44.5 & 47.1 \\
AgenTracer & 82.4 & 93.2 & 44.8 & 48.2 \\
\midrule
Ours (SFT) & 84.2 & \textbf{93.4} & 46.1 & 47.8 \\
\textbf{Ours (SFT + GRPO)} & \textbf{84.8} & 93.0 & \textbf{46.3} & \textbf{48.4} \\
\bottomrule
\end{tabular}
\end{table}

\textbf{Main results.}
In \cref{tab:self-reflection-main}, we present the Pass@1 accuracy on MATH500, GSM8K, KodCode, and MBPP.
\textbf{Ours (SFT + GRPO)} achieves a score of $\textbf{84.8}\%$ on MATH500, outperforming vanilla prompting, Self-reflection, and \textbf{AgenTracer}. Moreover, it achieves the best performance among all methods on the more challenging agentic tasks.
Building upon the SFT stage, GRPO further improves performance on MATH500, KodCode and MBPP.
On GSM8K, the baselines already have quite good performance over $92.6\%$. Thus, the absolute improvements by our methods are small near the ceiling.
In sum, SFT on our synthetic data delivers the main gain over vanilla prompting, Self-reflection, and \textbf{AgenTracer} on MATH500 and coding, and SFT already captures most of it. Applying GRPO adds a modest extra bump where accuracy still has headroom.

\textbf{Attribution analysis.}
To evaluate whether the learned attributions support targeted repair and generalize beyond the training domain, we train the attributor on math tasks and test it on Qwen3-8B failures from MATH500, GSM8K, AIME 2024, and LogiQA~2.0, with $83$/$84$/$22$/$441$ failed items, respectively.
Using \method{}'s paired-rollout labels as gold root steps, \cref{tab:attribution-rescue} shows the correction success rate -- a higher value indicates more accurate failure attribution. We can see that \textbf{Ours (SFT + GRPO)} consistently outperforms Self-reflection: $\textbf{19.3}\%$ vs.\ $11.9\%$ on MATH500, $\textbf{14.3}\%$ vs.\ $7.1\%$ on GSM8K, $\textbf{22.7}\%$ vs.\ $9.1\%$ on AIME 2024, and $\textbf{6.6}\%$ vs.\ $1.2\%$ on LogiQA~2.0.
These results show that hypergraph-verified attribution provides more actionable repair signals than self-reflection and transfers well from mathematical reasoning to out-of-domain logical reasoning.

\begin{table}[h]
\centering
\caption{Correction success rate (\%) on samples failed at first shot: the fraction of problems that Qwen3-8B initially fails and then solves after one attributor-guided repair (Pass@1 by default; Pass@16 on AIME 2024).}
\label{tab:attribution-rescue}
\vspace{-0.1in}
\small
\begin{tabular}{@{}lcccc@{}}
\toprule
\textbf{Method} & \textbf{MATH500} & \textbf{GSM8K} & \textbf{AIME 2024} & \textbf{LogiQA 2.0} \\
\midrule
Self-reflection & 11.9 & 7.1 & 9.1 & 1.2 \\
\textbf{Ours (SFT + GRPO)} & \textbf{19.3} & \textbf{14.3} & \textbf{22.7} & \textbf{6.6} \\
\bottomrule
\end{tabular}
\end{table}

\begin{table}[h]
\centering
\caption{MATH500 accuracy using different inference models. Our attributor generally outperforms baselines on different models.}
\label{tab:cross-model}
\vspace{-0.1in}
\small
\begin{tabular}{lcccc}
\toprule
\textbf{Method} & \textbf{Qwen3-8B} & \textbf{Qwen3-14B} & \textbf{Gemma3-4B} & \textbf{Llama-3.1-8B} \\
\midrule
Vanilla prompting & 79.4 & 81.0 & 20.4 & 46.8 \\
Self-reflection & 81.8 & 84.4 & \textbf{21.4} & 49.4 \\
\midrule
Ours (SFT) & 84.2 & \textbf{85.0} & 21.0 & 51.2 \\
Ours (SFT + GRPO) & \textbf{84.8} & 84.2 & 21.0 & \textbf{52.4} \\
\bottomrule
\end{tabular}
\end{table}

\textbf{Consistent gains across models.}
\cref{tab:cross-model} varies only the backbone models used for reflective reasoning, while using the same attributor checkpoint and prompt template.
Our attributor methods improve over vanilla prompting on backbones.
On Gemma3-4B, all methods performs poorly, near $20$--$21\%$. Considering the base model can only achieve 20\% accuracy, which is the lowest among models, we conjectures that attribution helps only when the base model can execute the repaired reasoning.

\section{Conclusion}

In this paper, we introduced \emph{Hypergraph-based Paired Failure Attribution} (\method{}), a framework that localizes the root causes of LLM reasoning failures by contrasting successful and failed trajectories and modeling their dependencies with a hypergraph. Paired trajectories narrow counterfactual verification to meaningful divergences, while hypergraph-guided search captures non-linear interactions that flat step sequences miss. Across mathematical reasoning, coding, and privacy-sensitive tasks, \method{} achieves a better localization cost--success trade-off than stepwise counterfactual and retrospective baselines, with the largest gains on long and heterogeneous traces. The verified attributions further provide scalable supervision for training a lightweight attributor, which improves reflective reasoning from a single trace on repair-capable backbones without modifying the underlying reasoning model. These results demonstrate that structured root cause feedback offers a practical path toward more reliable and scalable reflection.

\section*{Statements}
\label{sec:statements}

\noindent\textbf{Ethical Statement.}
\label{sec:ethical_statement}
This work develops failure-attribution methods to improve the reliability and self-correction of LLM reasoning. Such methods are dual-use and could be repurposed to refine undesirable behavior; however, our study is limited to controlled experiments on public math, coding, and privacy-preserving benchmarks and provides no guidance for harmful deployment.

\noindent\textbf{Reproducibility Statement.}
\label{sec:reproducibility_statement}
We report the models, data construction, hardware, and evaluation protocols in the main text and appendix. Code is available at \url{https://github.com/costa-nus/COLM26_HPFA} to support verification and extension of our results.

\noindent\textbf{Disclosure of LLM Use in Paper Preparation.}
\label{sec:llm_disclosure}
We used LLMs for language editing and preliminary literature review. The authors checked and revised all generated content, manually verified every citation, and take full responsibility for the paper.

\section*{Acknowledgments}

We appreciate the constructive feedback from reviewers in preparing the work. 
This research/project is partially supported by the National University of Singapore under the NUS Start-Up Grants (FY2026). OK's work was also carried out as part of the CYBERUS Erasmus Mundus Master's programme.
\bibliography{main}
\bibliographystyle{colm2026_conference}

\clearpage
\appendix

\section{Reliability of Hypergraph Construction}
\label{app:hypergraph-validation}

\noindent\textbf{Validation targets.}
We separately evaluate atomic decomposition, cross-trajectory merge decisions, and hyperedge extraction, followed by an end-to-end review of the finished hypergraphs.

\noindent\textbf{Cross-model agreement.}
On 200 problems, we compare the Qwen3-8B builder with three independent comparison models.
Each model re-runs the same construction stage on the same input.
An LLM alignment judge then matches units that express the same operation or dependency, allowing for wording differences.
The units are atomic steps in stage~1, cross-trajectory merge decisions in stage~2, and hyperedges in stage~3.
We aggregate matched units across problems and report micro-F1 agreement:
\[
100 \cdot \frac{2M}{|\mathrm{builder}| + |\mathrm{comparison}|},
\]
where $M$ is the number of matched builder--comparison unit pairs.
\Cref{tab:hypergraph-validation} reports the resulting agreement.

\noindent\textbf{Human review.}
Cross-model agreement measures construction stability rather than correctness.
To complement it, two human annotators independently review 50 finished hypergraphs for the end-to-end reasonableness of their atomic steps, cross-trajectory merges, and dependencies.
A graph passes only when both annotators judge it reasonable; 44 of 50 graphs pass under this strict criterion, yielding an 88\% pass rate.

\noindent\textbf{Conclusion.}
Agreement ranges from 76.3\% to 92.2\% across stages and comparison models.
The semantic-equivalence stage has the lowest agreement, consistent with the ambiguity of deciding when differently worded steps should be merged.
Together with the strict 88\% human pass rate, these results support the stability of the construction while leaving semantic alignment as its main source of uncertainty.

\begin{table}[H]
\centering
\caption{Per-stage micro-F1 agreement (\%) between the Qwen3-8B hypergraph builder and three independent comparison models on 200 problems.}
\label{tab:hypergraph-validation}
\small
\setlength{\tabcolsep}{7pt}
\begin{tabular}{lccc}
\toprule
Stage (builder = Qwen3-8B) & vs Claude Opus 4.7 & vs Gemini 3.5 Flash & vs GPT-5.5 \\
\midrule
Atomic decomposition & 88.6 & 92.2 & 84.9 \\
Semantic-equivalence judgment & 76.3 & 83.1 & 82.0 \\
Hyperedge extraction & 89.6 & 85.7 & 84.3 \\
\bottomrule
\end{tabular}
\end{table}

\section{Additional Attribution Analysis}
\label{app:attributor-analysis}
\label{sec:app-attribution-confusion}

To further analyze the gains of the trained attributor over reflection in \cref{sec:exp-attributor}, we decompose the accuracy change on the same 500 MATH500 problems.
\Cref{tab:attribution-confusion} cross-tabulates each problem's outcome before reflection (direct reasoning; rows) and after one reflection pass guided by each method (columns).
The diagonal cells are problems whose outcomes remain unchanged after reflection.
The upper-right cell $b$ counts \emph{regressions} from correct to incorrect, while the lower-left cell $c$ counts \emph{recoveries} from incorrect to correct.
Their difference, $\delta=c-b$, is the net number of rescued problems, corresponding to an absolute accuracy gain of $100\delta/500$ percentage points over direct reasoning.
To test whether recoveries and regressions are symmetric, we use the asymptotic McNemar test without continuity correction~\citep{mcnemar1947note}: for $b+c>0$, $\chi^2=(c-b)^2/(b+c)$ and the reported $p$-value is the upper-tail probability $\Pr(\chi^2_1\geq\chi^2)$.

\begin{table}[H]
\centering
\caption{Outcome transitions from direct reasoning to one reflection pass on the same 500 MATH500 problems.}
\label{tab:attribution-confusion}
\small
\begin{tabular}{lcccccc}
\toprule
\multirow{2}{*}{\textbf{Direct reasoning}} & \multicolumn{2}{c}{\textbf{Self-reflection}} & \multicolumn{2}{c}{\textbf{Ours (SFT)}} & \multicolumn{2}{c}{\textbf{Ours (SFT + GRPO)}} \\
 & Correct & Incorrect & Correct & Incorrect & Correct & Incorrect \\
\midrule
Correct   & 370 & 27 & 360 & 37 & 355 & 42 \\
Incorrect & 39  & 64 & 61  & 42 & 69  & 34 \\
\bottomrule
\end{tabular}
\end{table}

\begin{table}[H]
\centering
\caption{Regressions, recoveries, net rescues, and McNemar test results derived from \cref{tab:attribution-confusion}.}
\label{tab:attribution-mcnemar}
\small
\begin{tabular}{lcccc}
\toprule
\textbf{Method} & \textbf{Regressions $b$} & \textbf{Recoveries $c$} & $\delta=c-b$ & \textbf{$p$-value} \\
\midrule
Self-reflection & 27 & 39 & 12 & 0.140 \\
Ours (SFT) & 37 & 61 & 24 & 0.015 \\
Ours (SFT + GRPO) & 42 & 69 & 27 & 0.010 \\
\bottomrule
\end{tabular}
\end{table}

\noindent\textbf{Results.}
Self-reflection yields 12 net rescues, corresponding to a $2.4$-point gain over direct reasoning, but its recovery--regression asymmetry is not significant ($p=0.140$).
Training the attributor increases the net rescues to 24 with SFT and 27 with SFT + GRPO, exactly matching the $4.8$- and $5.4$-point gains in \cref{tab:self-reflection-main}; both asymmetries are significant ($p=0.015$ and $0.010$).
Relative to self-reflection, the trained attributors therefore add 12 and 15 net rescues, or $2.4$ and $3.0$ accuracy points.
Although training also introduces more regressions, its larger increase in recoveries produces a stronger net benefit, indicating that the learned attribution signal makes reflection more effective at repairing initially failed solutions.

\section{Privacy Scenario Details}
\label{sec:app-privacy-localization}

This appendix details the privacy scenario introduced in \cref{sec:exp-localization}, where failures caused by privacy-preserving de-identification must be distinguished from errors in the LLM's reasoning.

\noindent\textbf{Dataset and experimental setup.}
We construct the privacy slice from CoQA conversational questions~\citep{reddy2019coqa}.
We retain 108 cases in which the model fails on the de-identified input and the privacy checks below are satisfied; attribution then determines whether each failure is scrub-induced or reasoning-induced.
All LLM-based components use GPT-4o, while PrivateAI is used only for de-identification.

\noindent\textbf{Evaluation procedure.}
For each instance, we apply the following four steps:
\begin{enumerate}[leftmargin=*,itemsep=2pt,parsep=0pt,topsep=4pt]
  \item \textbf{De-identify.} PrivateAI replaces sensitive spans in the original query \(q\) and its context with typed placeholders, producing the scrubbed view \(\tilde{q}\).
  \item \textbf{Generate paired trajectories.} In controlled evaluation, we obtain a reference trajectory \(\cT^{+}\) from \(q\) and a failed privacy-constrained trajectory \(\cT^{-}\) from \(\tilde{q}\); only \(\tilde{q}\) represents the input available in deployment.
  \item \textbf{Localize and verify.} \method{} aligns the two trajectories, constructs their dependency hypergraph, and counterfactually verifies the earliest divergent step.
  \item \textbf{Attribute the cause.} We classify the verified step using the taxonomy below. Scrub-induced failures motivate changes to redaction or placeholder policies, whereas reasoning failures motivate changes to prompting or decoding.
\end{enumerate}

\noindent\textbf{Privacy constraints.}
We apply two checks.
First, the scrubber must only replace content rather than introduce new factual text.
Let \(\mathrm{strip}(\tilde{q})\) remove placeholder tokens from the scrubbed query; after whitespace normalization, we require
\begin{equation}
\label{eq:privacy-substring}
  \mathrm{strip}(\tilde{q}) \;\in\; \mathrm{Substr}(q).
\end{equation}
Here \(\mathrm{Substr}(q)\) is the set of normalized contiguous substrings of \(q\).
Second, the generated answer must not reproduce any private field from \(q\) verbatim.
We evaluate correctness against the CoQA reference answer, and a counterfactual repair counts as successful only if it is correct and passes the no-leakage check.

\noindent\textbf{Taxonomy of scrub-related and reasoning failures.}
To make the attribution actionable, we classify the localized cause using the five categories in \cref{tab:privacy-error-taxonomy}.
The first, second, fourth, and fifth categories indicate failures introduced by the privacy transformation; \emph{Reasoning Failure} is reserved for cases in which the scrubbed input still contains all evidence needed to answer correctly.

\begin{table}[H]
\centering
\caption{Error taxonomy for the privacy scenario.}
\label{tab:privacy-error-taxonomy}
\small
\begin{tabular}{@{}p{0.18\linewidth}p{0.76\linewidth}@{}}
\toprule
\textbf{Category} & \textbf{Localized cause} \\
\midrule
Cross-reference &
The same entity is replaced by different placeholders, preventing the model from recognizing that the mentions co-refer.
\\
\midrule
Embraced Scrubbing &
The target answer is hidden within or split across scrubbed spans, making it unavailable for extraction or reconstruction.
\\
\midrule
Reasoning Failure &
All necessary evidence is preserved in \(\tilde{q}\), but the model makes an incorrect inference.
\\
\midrule
Unresolved Coreference &
Redaction prevents a pronoun or generic expression from being linked to its antecedent.
\\
\midrule
Scrubbing Divergence &
An entity is represented inconsistently across the question and context, such as a placeholder in one view and plain text or a different label in another.
\\
\bottomrule
\end{tabular}
\end{table}

\noindent\textbf{Takeaway.}
This privacy scenario extends the localization evaluation in \cref{sec:exp-localization} beyond ordinary reasoning errors.
By identifying both the failed reasoning step and whether its cause lies in scrubbing, \method{} provides a verified target for repair without treating every failure as a decoder problem.

\section{Prompt Templates}
\label{app:prompts_en}

Table~\ref{tab:prompt_families_en} summarizes the role, inputs, and outputs of each prompt family; full templates follow.

\begin{table}[h]
\centering
\caption{Prompt families used in our experiments.}
\label{tab:prompt_families_en}
\small
\begin{tabular}{p{2.1cm}p{3.8cm}p{4.4cm}p{2.6cm}}
\toprule
\textbf{Prompt family} & \textbf{Role} & \textbf{Inputs} & \textbf{Outputs} \\
\midrule
Trajectory sampling & One solution per task per call; repeatable for multiple trajectories & Task text; format constraints from the template & Full generations that follow the template \\
Paired attribution & Compare reference vs.\ failing trace; locate the earliest erroneous step on the failure chain & Query; reference and failing traces; answer or score summary; optional prior hints & Structured description of the root-cause step \\
Negative-only attribution & Without a gold reference, locate the earliest error on the failure chain (same output format as paired) & Query; failing trace; answer or score summary; optional prior hints & Same \texttt{root\_cause} tag structure as paired (\texttt{step\_id}, \texttt{step\_content}) \\
Hypergraph construction & Atomize reasoning steps and build a dependency hypergraph for parsing & Structured instance with query and pos./neg.\ attempts; optional few-shot and target instance & Structured steps and dependency relations \\
Re-reasoning & Rewrite a solution from root-cause diagnosis without external gold references & Query; failing trace; root-cause text & Revised full generation under tag constraints \\
\bottomrule
\end{tabular}
\end{table}

\phantomsection\label{app:sampling_prompts_en}

\begin{promptcard}{Trajectory sampling / Code}
\phantomsection\label{app:sampling_code_en}
\begin{lstlisting}[style=promptsnippet]
"""/no_think You are an expert Python programmer. Solve the following programming task with a correct, runnable Python solution.

### Task
{question}

### Output format (required)
1. You may briefly reason outside the tags if needed.
2. Put your **final, complete solution** (the full Python code the grader should run) inside **exactly one** pair of tags:
<code>
# your complete code here
</code>
3. Use the literal tag names `<code>` and `</code>` (lowercase). Do not use markdown fences instead of these tags for the final code.
4. The code inside `<code>...</code>` must be self-contained as much as the task allows.
5. If the task names a function or class (e.g. `def maximum_product_of_three(...)`), you **must** define that exact name in your code so `from solution import <that name>` succeeds; do not rename, omit, or only define helpers with different names.

Your answer:
"""
\end{lstlisting}
\end{promptcard}

\phantomsection\label{app:rca_prompts_en}

\begin{promptcard}{Paired attribution (success / failure) / Math}
\phantomsection\label{app:paired_rca_prompt_en}
\begin{lstlisting}[style=promptsnippet]
"""/no_think You are given a math problem and two solution attempts: one correct (positive) and one wrong (negative). The **failure solution** is written **step-by-step** (e.g. "Step 1", numbered items, or similar).

**Problem (query):**
{query}

**Groundtruth solution (correct, answer = {positive_answer}):**
{positive_response}

**Failure solution (wrong, answer = {negative_answer}):**
{negative_response}
{prev_hint}

Compare the two reasoning chains. In the **failure solution only**, locate the **single step** where the reasoning first goes wrong (the earliest step whose mistake leads to the wrong final answer).

Put **only** the following inside one pair of tags, nothing outside the tags (no preamble, no analysis):

<root_cause>
step_id: <the step label or number exactly as in the failure solution>
step_content: <the full text of that step, copied verbatim from the failure solution>
</root_cause>"""
\end{lstlisting}
\end{promptcard}

\begin{promptcard}{Negative-only attribution / Math}
\phantomsection\label{app:failed_only_rca_en}
\begin{lstlisting}[style=promptsnippet]
"""/no_think You are given a math problem and **one wrong solution attempt** (negative) only. **No correct reference solution is provided.** The **failure solution** is written **step-by-step** (e.g. "Step 1", numbered items, or similar).

**Problem (query):**
{query}

**Failure solution (wrong, answer = {negative_answer}):**
{negative_response}
{prev_hint}

In the **failure solution only**, locate the **single step** where the reasoning first goes wrong (the earliest step whose mistake leads to the wrong final answer). Infer the error using the problem statement and the failure trace alone.

Put **only** the following inside one pair of tags, nothing outside the tags (no preamble, no analysis):

<root_cause>
step_id: <the step label or number exactly as in the failure solution>
step_content: <the full text of that step, copied verbatim from the failure solution>
</root_cause>"""
\end{lstlisting}
\end{promptcard}

\begin{promptcard}{Hypergraph construction / Code}
\phantomsection\label{app:hypergraph_prompt_en}
\begin{lstlisting}[style=promptsnippet]
"""You are an expert in structured analysis of **programming solution attempts** (natural-language reasoning plus Python code).

Your task is to analyze **two** attempts for the **same** coding problem, a **positive** attempt (passes tests) and a **negative** attempt (fails), and:

1. **Atomize steps**: Decompose each side into minimal units. **positive** maps to trajectory label **T1**, **negative** to **T2** in `trajectoryies_id`.
2. **Build a hypergraph**: Within-side dependencies, then **merge** nodes only when **both** **positive.response** and **negative.response** **explicitly** contain the same step (same code intent). Do not merge on guesswork.
3. **Output** in the standard format below.

---

### Hypergraph structure notes

- **T1** <-> **positive** (passing). **T2** <-> **negative** (failing).
- Merge only with strict textual evidence in both responses.
- `trajectoryies_id` uses "T1" and/or "T2" accordingly.

---

Input format (**minimal adversarial shape**; **no `trajectories` key**; **no** `gt_solution` / `test` / `test_info`):
```json
{
  "id": "string",
  "query": "The programming problem statement",
  "positive": {
    "response": "Full model output for the passing attempt",
    "code": "Program that passes tests"
  },
  "negative": {
    "response": "Full model output for the failing attempt",
    "code": "Program that fails tests"
  }
}
```

---

Your output format:
```json
{
  "steps": [
    {
      "node_id": "S1",
      "trajectoryies_id": ["T1"],
      "text": "Content of one atomic step"
    }
  ],
  "dependencies": [
    {
      "from": ["S1"],
      "to": ["S2"]
    }
  ]
}
```

Field descriptions:
1. node_id: Unique step id. Merged steps use trajectoryies_id ["T1","T2"] when both sides explicitly share the step.
2. trajectoryies_id: **T1** always labels steps that come from the **positive** trajectory (the passing attempt: **positive.response** / **positive.code**). **T2** always labels steps that come from the **negative** trajectory (the failing attempt: **negative.response** / **negative.code**). Use ["T1"], ["T2"], or ["T1","T2"] according to which side(s) the step belongs to.
3. text: Clear programming/reasoning step description.
4. dependencies: As before.

---

Below is a few-shot example:
few_shot_example:
${few_shot_example}

---

Your actual input:
${real_input}

"""
\end{lstlisting}
\end{promptcard}

\begin{promptcard}{Re-reasoning / Code}
\phantomsection\label{app:rereason_prompt_en}
\begin{lstlisting}[style=promptsnippet]
"""/no_think You are given a programming task. A previous solution was wrong. The root cause analysis below points to the earliest mistake. **You are not given any reference or gold solution**. Write a **new** complete Python program that fixes the issue implied by the root cause and correctly solves the task on your own.

**Task:**
{query}

**Previous wrong attempt (for context only; do not repeat its errors):**
{negative_response}

**Root cause (what went wrong):**
{root_cause}

Provide a brief plan if needed, then put the **full final program** in **exactly one** pair of tags:
<code>
# your complete solution
</code>
Use lowercase `<code>` / `</code>`. Your answer:"""
\end{lstlisting}
\end{promptcard}

\end{document}